\documentclass[letterpaper, 10 pt, conference]{ieeeconf}
\IEEEoverridecommandlockouts                              

\IEEEoverridecommandlockouts                              

\usepackage{float}
\usepackage{booktabs}

\usepackage{graphicx}
\usepackage{caption}
\usepackage{subcaption}
\usepackage{xcolor}
\usepackage{multirow}
\usepackage[colorlinks,linkcolor=blue,citecolor=blue,urlcolor=blue]{hyperref}
\hypersetup{hidelinks,
	colorlinks=true,
	allcolors=black,
	pdfstartview=Fit,
	breaklinks=true}

\usepackage{multirow}

\makeatletter
{\catcode`\:=12 \catcode`\-=12 \catcode`\>=12 \catcode`\<=12 %
 \gdef\codeof#1{\expandafter\codeof@\meaning#1<-:}%
 \gdef\codeof@#1:->#2<-:{#2}}
\def\bibwrap#1#2{%
\bibitem{#1}#2%
\if@filesw{\def\next{{#2}}%
           \immediate\write\@auxout{\string\long@bibcite{#1}\codeof\next}}%
\fi}
\def\long@bibcite{\@newl@bel y} 
\def\long@cite#1{\csname y@#1\endcsname
    \immediate\write\@auxout{\string\citation{#1}}}
\def\longcite#1{[\long@cite{#1}]}
\makeatother

\title{\LARGE \bf
AVD2: Accident Video Diffusion for Accident Video Description
}

\author{Cheng Li$^{1,2,*}$, Keyuan Zhou$^{1,3,*}$, Tong Liu$^{1,4,*}$, Yu Wang$^{1,5,*}$, Mingqiao Zhuang$^{6}$,\\
Huan-ang Gao$^{1}$, Bu Jin$^{1}$ and Hao Zhao$^{1,7,8,\dagger}$
\thanks{$^{\dagger}$The corresponding author.}
\thanks{$^{*}$Indicates equal contribution.}
\thanks{$^{1}$Institute for AI Industry Research (AIR), Tsinghua University.}%
\thanks{$^{2}$Academy of Interdisciplinary Studies, the Hong Kong University of Science and Technology.}%
\thanks{$^{3}$College of Communication Engineering, Jilin University.}%
\thanks{$^{4}$School of Cyber Science and Engineering, Nanjing University of Science and Technology.}%
\thanks{$^{5}$School of Automation, Beijing Institute of Technology.}%
\thanks{$^{6}$College of Foreign Language and Literature, Fudan University.}%
\thanks{$^{7}$Beijing Academy of Artificial Intelligence (BAAI).}%
\thanks{$^{8}$Lightwheel AI.}%
}

\let\oldtwocolumn\twocolumn
\renewcommand\twocolumn[1][]{%
    \oldtwocolumn[{#1}{
    \begin{center}
    \vspace{-1em}
        \includegraphics[width=0.95\linewidth]{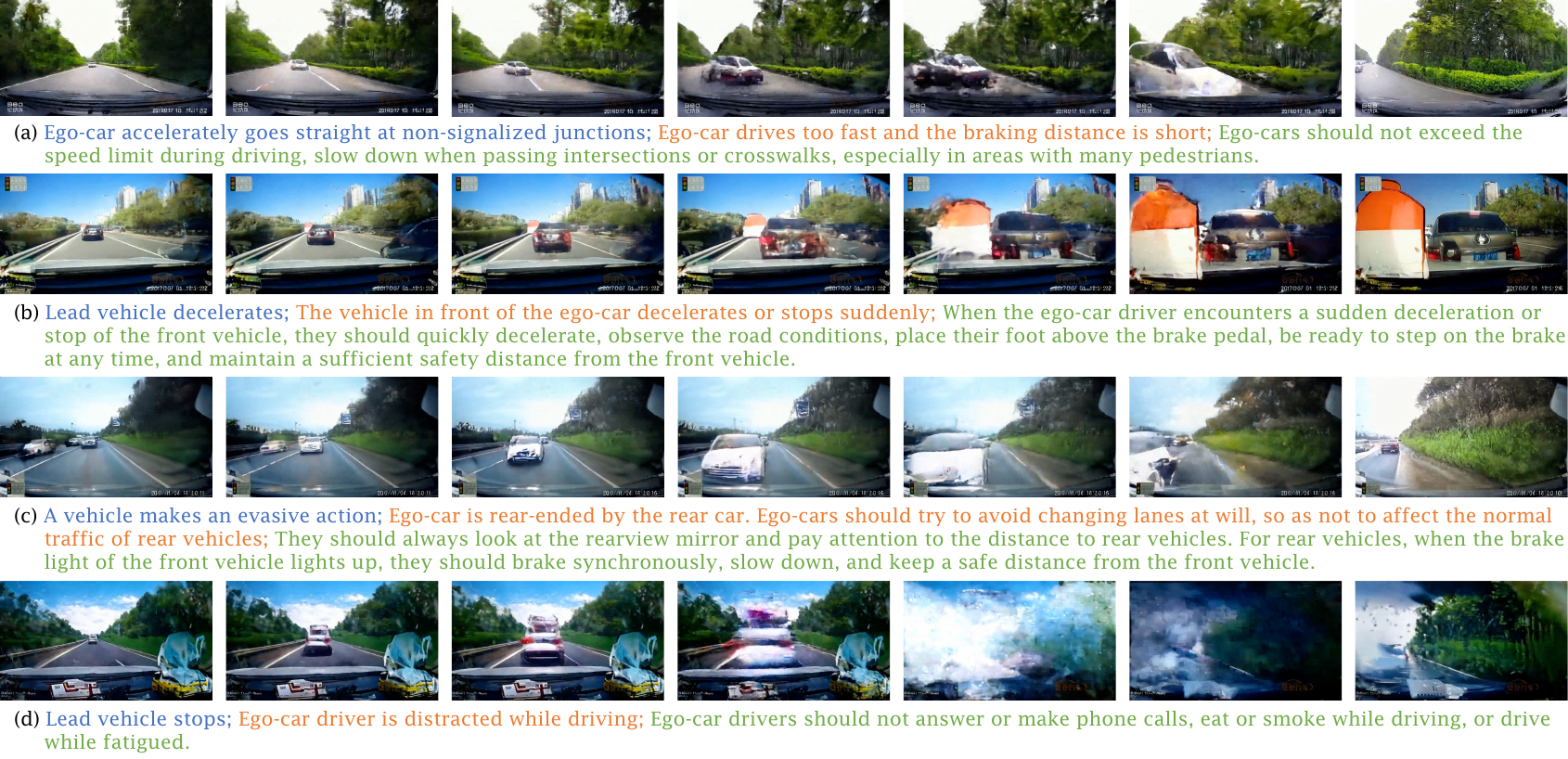}
        \vspace{-0.3em}\captionof{figure} {\textbf{Accident Video Frames from our contributed EMM-AU dataset.} We are the first to generate realistic traffic accident scenarios, improving natural language descriptions and reasoning for autonomous driving.}
        \label{fig:teaser}
    \end{center}
    }]
}

\begin{document}

\maketitle

\begin{abstract}
Traffic accidents present complex challenges for autonomous driving, often featuring unpredictable scenarios that hinder accurate system interpretation and responses. 
Nonetheless, prevailing methodologies fall short in elucidating the causes of accidents and proposing preventive measures due to the paucity of training data specific to accident scenarios. In this work, we introduce AVD2 (Accident Video Diffusion for Accident Video Description), a novel framework that enhances accident scene understanding by generating accident videos that aligned with detailed natural language descriptions and reasoning, resulting in the contributed EMM-AU (Enhanced Multi-Modal Accident Video Understanding) dataset. Empirical results reveal that the integration of the EMM-AU dataset establishes state-of-the-art performance across both automated metrics and human evaluations, markedly advancing the domains of accident analysis and prevention. Project resources are available at \href{https://an-answer-tree.github.io}{https://an-answer-tree.github.io}
\end{abstract}

\section{Introduction}
End-to-end autonomous driving systems have emerged as a leading approach in the development of intelligent vehicles, primarily due to their ability to streamline the decision-making process by directly mapping sensor inputs to control signals, minimizing accumulative errors in modular designs \cite{chen2024end,korhonen2012peak}.
These systems integrate perception, planning, and control into a cohesive unit, enabling a more fluid and efficient response to driving conditions. However, despite their potential, the robustness of these systems remains a significant challenge, particularly when faced with accident scenarios \cite{jin2023adapt,marcu2024lingoqa}. Accidents are not only costly but also present complex and unpredictable situations that can severely test the limits of autonomous driving technology. Before these systems can be deployed on a global scale, it is crucial to understand the intricacies of traffic accidents and develop effective interpretation and reasoning mechanisms to explain and prevent them. This understanding is essential for enhancing the safety and reliability of autonomous vehicles, ultimately fostering greater public trust in this transformative technology.


When an accident occurs, it is imperative to answer key questions: what caused the accident, what were the contributing factors, and how can similar incidents be prevented? Existing approaches, such as Video Question Answering~\cite{atakishiyev2023explaining,xu2021sutd,qian2024nuscenes}, have attempted to address these questions by formulating accident reasoning within a question-answering framework. The existing method for modeling videos for QA involves creating neural architectures where each sub-system is crafted for a specific, custom purpose or a particular type of data. This specialized design often makes these manually created architectures less effective when faced with changes in data modality, varying video lengths, or different types of questions~\cite{le2020hierarchical}.
Moreover, these methods often struggle with the complexity and unpredictability of real-world scenarios, failing to offer effective prevention strategies.

To address these limitations, we propose a novel framework called AVD2 (Accident Video Diffusion for Accident Video Description), which contains a video generation pipeline and an accident analysis system. The generation pipeline of AVD2 aims to enhance the understanding of accident scenes by generating accident videos that are meticulously aligned with detailed description of the accident, including the caption of the accident video, an analysis of the cause, and preventive measures recommendations (avoidance).
By providing detailed explanation and actionable avoidance, AVD2 improves the accuracy and relevance of video generation, resulting in our contributed EMM-AU dataset, which incorporates a large number of newly generated accident videos, provides a robust foundation for developing more accurate and reliable accident analysis methods.

Built upon EMM-AU, we propose a novel architecture for accident video understanding that concurrently generates descriptions of the input accident videos and provides reasoning on effective avoidance strategies. Central to this framework is the employment of SCST (self-critical sequence training), which emphasizes enhanced focus on contextual relevance. Within the AVD2 framework, SCST substantially enhances the accuracy and contextual alignment of captions with visual content. This improvement is critical for capturing the intricate details of complex driving scenarios and increasing the reliability of autonomous driving systems.

Through extensive experiments, we demonstrate that our AVD2 framework addresses the challenge of accident reason answering by generating detailed actionable insights. This enhancement is crucial for increasing user trust and ensuring that autonomous vehicles can be safely integrated into real traffic environments.
Our contributions can be summarized as:
\begin{itemize}
\item We propose a novel framework, \textbf{AVD2} (Accident Video Diffusion for Accident Video Description), which integrates a video generation pipeline and an accident analysis system. This framework is designed to generate detailed accident videos aligned with comprehensive descriptions, including captions, cause analyses, and preventive measures.
\item Utilizing the AVD2 framework, we compile the Enhanced Multi-Modal Accident Video Understanding (EMM-AU) dataset. This extensive dataset, consisting of newly generated accident videos, provides a robust foundation for developing more precise and reliable methods for accident analysis and prevention.
\item Built upon the EMM-AU dataset, we develop an advanced architecture for accident video understanding that simultaneously generates descriptions and reasoning on effective avoidance strategies for the input accident videos.
\end{itemize}

\section{Related Work}

\subsection{Interpretability and Accident Analysis in Autonomous Driving}

Interpretability is crucial in autonomous driving systems for enhancing user trust and ensuring transparent decision-making. Traditional methods primarily leverage vision-based and LiDAR-based technologies. Vision-based systems employ visual attention mechanisms and segmentation maps to provide insights into vehicle behavior based on its surroundings~\cite{kim2017interpretable}. LiDAR-based systems generate three-dimensional maps and detect objects to improve spatial awareness and interaction understanding~\cite{zeng2019end}. However, these techniques often fail to deliver explanations in natural language, making it challenging for non-expert users to comprehend the vehicle's decisions.

To address this, Kim et al.~\cite{kim2018textual} introduced textual explanations, translating vehicle behaviors and decision-making processes into comprehensible language. Building on this, the Action-aware Driving Caption Transformer (ADAPT) framework~\cite{jin2023adapt} utilizes a transformer-based architecture to elucidate the ‘why’ and ‘how’ behind autonomous decisions. Our AVD2 framework advances this approach by integrating Self-critical Sequence Training (SCST)-based contextual analysis~\cite{bujimalla2020b}, further enhancing the clarity and accessibility of accident explanations. Recognizing the complexity of real-world traffic scenarios, existing methods, such as Video Question Answering (VQA) models, attempt to analyze accident causes by combining visual data with structured queries~\cite{xu2021sutd,liu2023cross}. However, these approaches often struggle to capture the complexity and variability of real-world traffic scenarios~\cite{le2020hierarchical}. By leveraging the strengths of ADAPT, AVD2 significantly enhances the precision and contextual relevance of accident explanations. AVD2 generates detailed descriptions, analyzes underlying causes, and provides actionable prevention strategies, setting a new standard for explainability and reliability in autonomous driving systems~\cite{nahata2021assessing,hwang2024safe,ding2024hint,li2023understanding}.

\begin{figure*}[htbp]
\centering
\includegraphics[width=0.8\linewidth]{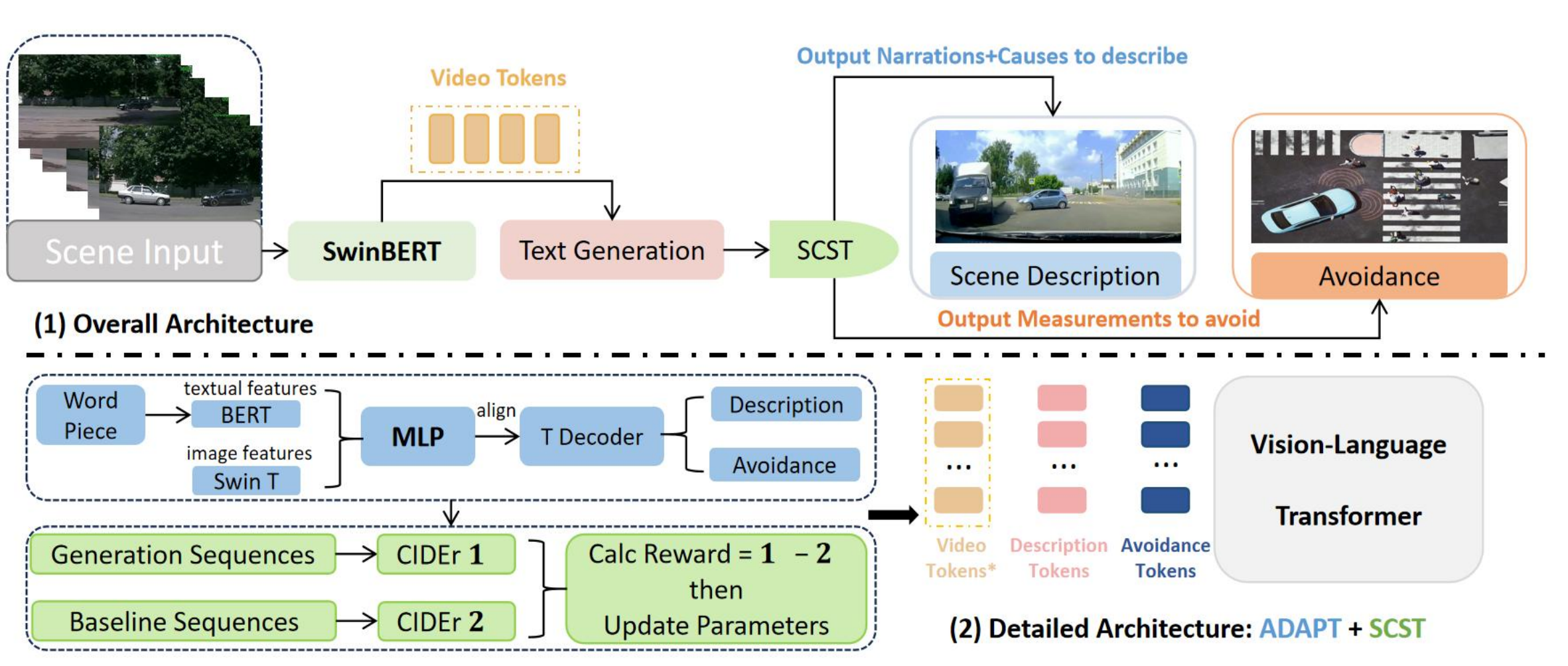} 
\caption{\textbf{The Framework Architecture of AVD2 system.} {\footnotesize The frame diagram demonstrates a visual language system for generating descriptions and obstacle avoidance cues from video. SwinBERT processes the video input, converts frames into video tags, and outputs descriptions and obstacle avoidance suggestions via a text generation module. The description includes the driving situation of the vehicle, and the obstacle avoidance section gives safety suggestions. Visual-language Transformer extracts text and image features and optimizes the generation with SCST.}}
\label{fig:method}
\vspace{-1em}
\end{figure*}

\subsection{Video Captioning for Autonomous Driving}

Video captioning converts video content into natural language by integrating computer vision and natural language processing. Early research focused on generating sentences with specific syntactic structures by filling recognized elements into fixed templates~\cite{inoue2023towards}, which lacked flexibility and richness. As the field progressed, sequence learning approaches were introduced, allowing for the generation of more natural sentences with flexible syntactic structures~\cite{venugopalan2015sequence}. Further developments included the use of object-level representations to capture fine-grained object-aware interactions in videos~\cite{gao2023retrieval}. More recently, Transformer-based models like SWINBERT have been introduced, employing sparse attention mechanisms to reduce redundant information and enhance the descriptive precision of video captions~\cite{wang2021end,lin2022swinbert,liu2023delving,zheng2024monoocc}.

However, current models struggle with accurately representing complex actions, especially in autonomous driving scenarios where capturing nuanced details is crucial. This limitation hampers their effectiveness in scenarios requiring detailed analysis, such as accident understanding and prevention. To address these challenges, we introduce the AVD2 framework, designed to enhance accident comprehension. AVD2 enhances the precision and contextual relevance of video captions, particularly in complex scenarios like traffic accidents, providing actionable insights for prevention.

\subsection{Video Generation}

Video generation utilizes computational techniques and models to produce video content through advanced methodologies ~\cite{vondrick2016generating,cai2018deep,huang2024vbench,chen2023pixart}. Key methods include image-to-video transformation \cite{hussain2017automatic,karras2023dreampose}, text-to-video synthesis \cite{karras2023dreampose}, and the reconfiguration of existing video materials into new segments~\cite{jiang2023text2performer}. Among these, text-to-video generation is particularly important in autonomous driving \cite{wen2024panacea,yuan2024magictime}, where creating realistic and diverse scenarios is essential for developing and testing system modules. Recent advancements in large language models (LLMs)
, have enabled the generation of highly detailed and realistic scenes~\cite{zhang2024ctrl, xu2024diffusion, gao2024scp, li2024fairdiff, gu2025text2street}.
 Techniques like scenario prompt engineering, scenario generation through LLMs, and iterative evaluation feedback have been employed to improve these models~\cite{chang2024llmscenario}.

Despite these efforts, generating accurate and contextually appropriate accident videos remains a challenge. Previous research has made strides in video generation, yet these models often struggle with the complexity and unpredictability of real-world traffic accidents fully~\cite{tulyakov2018mocogan,pang2021image}. Moreover, the application of text-to-video generation specifically for accident scenarios is still underexplored~\cite{denton2018stochastic}. To address these gaps, we propose a novel framework that leverages Open-Sora 1.2 to generate high-fidelity accident videos. By fine-tuning pre-trained models with MM-AU raw data, our approach achieves the first successful generation of detailed and realistic accident scenarios~\cite{gupta2018social}. This innovation sets a new benchmark in autonomous driving research, significantly improving the quality and diversity of training datasets and providing more robust evaluation and training tools for autonomous driving systems.

\section{Methodology}


\subsection{Augmentation and Enhancement of Video Dataset}
To facilitate the training of accident understanding models, we developed the \textbf{EMM-AU} (Enhanced MM-AU) dataset. Derived from the MM-AU dataset~\cite{fang2024abductive}, EMM-AU integrates advanced video generation \cite{zheng2024open,blattmann2023stable} and enhancement~\cite{wang2021real} techniques. It is the first dataset designed specifically for the Driving Accident Reasoning task that incorporates video generation models for data augmentation.

\subsubsection{The MM-AU Dataset}

The MM-AU dataset is the largest and most comprehensive ego-view multi-modal accident dataset to date, comprising 8,399 annotated videos, totaling 147 hours of curated content. These videos are primarily sourced from Ads-of-the-World~\cite{ads_of_the_world} (6,304 videos), with additional footage from the Cannes Lions International Festival of Creativity~\cite{cannes_lions}  (1,135 videos) and the Video-Ads dataset~\cite{hussain2017automatic} (960 videos).
To better represent driving accidents in our research, we restructured the MM-AU annotations. We merged the original “texts” and “causes” fields into a new “description” field while retaining the “avoidance” data from the original “measures” field.

\subsubsection{The introduction of EMM-AU Dataset}

To create EMM-AU, we leveraged Open-Sora 1.2 \cite{brooks2024video}, an advanced open-source text-to-video model capable of generating 720p, 16-second HD videos. Building on the official Open-Sora 1.2 pre-trained model, we fine-tuned it using the original MM-AU dataset, employing the “descriptions" and “avoidance" fields as prompts. We then conducted two stages of fine-tuning: the first stage involved training on the full MM-AU dataset to mimic the accident scene style, while the second stage focused on additional fine-tuning using 2,000 selected detailed accident scene videos from MM-AU dataset for better text alignment. After the training process, we run inference to produce 2000 new videos, which were then integrated into a new dataset, dubbed as EMM-AU. This augmentation process significantly enriched the dataset, adding greater depth and diversity \cite{zheng2024open,blattmann2023stable}. Figure 1 illustrates the incident frames of four exemplary videos generated by our fine-tuned model, utilizing prompts derived from the original MM-AU dataset annotations. Additionally, Figure 3 presents two examples generated by the original Open-Sora model, employing the same prompts as those in Figure 1(a).

\begin{figure}[htbp]
    \centering
    \includegraphics[width=\linewidth]{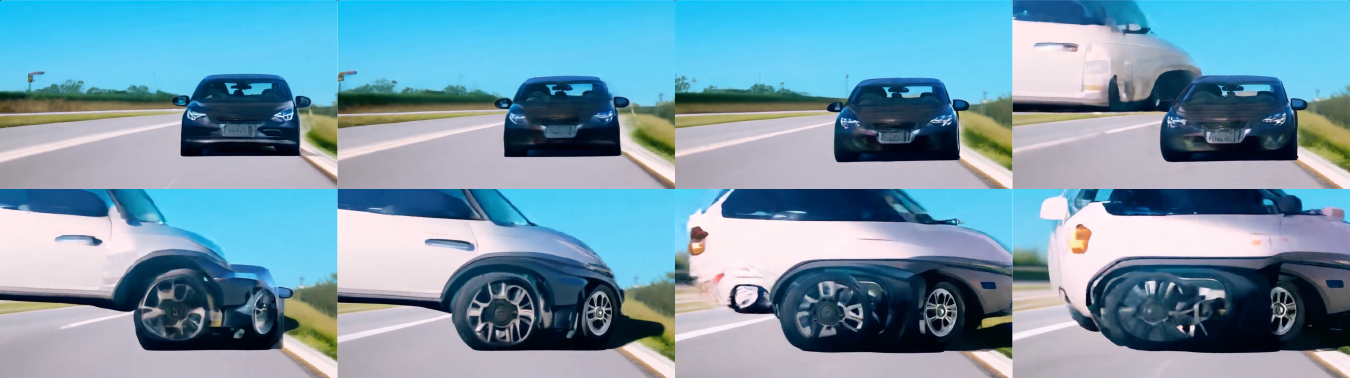} 
    \caption{\textbf{The Incident Frames of Video Generated by original Open-Sora Model.} The prompts used the original MM-AU dataset annotations.}
    \vspace{-0.8em}
\end{figure}

\subsubsection{Super-Resolution for Video Quality Enhancement}
 To enhance video quality, we employed the RRDBNet model within the Real-ESRGAN framework~\cite{wang2021real,kang2023scaling}, chosen for its ability to upscale low-resolution images by four times while preserving details and reducing artifacts~\cite{wang2018esrgan,dosovitskiy2016generating}. The process begins with initializing the RRDBNet model, configured with 23 residual blocks to optimize super-resolution tasks. Each video frame is processed individually, ensuring the upscaled output remains within a 4K resolution limit~\cite{zhang2023blind,kang2023scaling} , balancing visual quality with computational efficiency.

These upscaled frames were then resized to meet the target resolution, ensuring high-definition quality for further processing. The enhanced videos serve as inputs for the AVD2 framework to evaluate the effectiveness of data augmentation strategies using the fine-tuned Open-Sora 1.2 model~\cite{yu2023language,li2024carbon}.

\subsection{AVD2 System}        
\subsubsection{Overview}

We present the AVD2 system which leverages the strengths of both ADAPT and SCST to enhance the accuracy and relevance of captions for autonomous driving scenarios. It produces captions responsive to dynamic driving actions by combining visual and contextual data from video sequences~\cite{chen2024end,jin2023adapt}. For further enhancement, we incorporated the SCST mechanism and apply reinforcement learning to optimize sequence generation, directly targeting evaluation metrics such as CIDEr~\cite{rennie2017self,xu2019multi,luo2018discriminability}, which ensures captions not only match the context but also align closely with human-assessed quality metrics, making it highly promising for practical applications in accident scene understanding. The architecture of the AVD2 framework is shown in Figure 2.

\subsubsection{Action-aware Driving Caption Transformer}

The ADAPT framework aims to generate contextually relevant captions for autonomous driving scenarios by combining the functionality of the Swin Transformer and BERT architectures. Swin Transformer serves as the visual backbone, and through its hierarchical structure and shift-window approach~\cite{jin2023adapt} effectively captures the local and global features in video frames. This allows the model to understand the intricate details and broader context of the driving environment. Simultaneously, BERT processes the textual data, ensuring that the captions generated are both semantically rich and contextually appropriate. ADAPT aligns visual and textual information to produce accurate, meaningful descriptions of vehicle actions in complex driving scenarios.

Building on the ADAPT framework, we developed the AVD2 to specifically address accident scenario understanding. While ADAPT focuses on general driving captions, AVD2 adapts this approach by concentrating on two main tasks: describing the actions leading to an accident (action; cause) and proposing preventive measures (avoidances). AVD2 leverages the strong visual and textual alignment capabilities of ADAPT, but reorients them to focus on generating detailed narratives of accident events and actionable insights to prevent similar incidents. This adaptation ensures that AVD2 provides not only a descriptive understanding of accidents but also practical recommendations for improving safety.
\subsubsection{Self Critical Sequence Training}

SCST is a reinforcement learning method that optimizes sequence generation models by aligning them with evaluation metrics like CIDEr, BLEU, and METEOR~\cite{rennie2017self}. Unlike traditional training, which maximizes the likelihood of generating sequences based on ground truth, SCST uses a reward mechanism to encourage higher metric scores~\cite{bujimalla2020b,luo2020better}.

In the AVD2 framework, SCST enhances ADAPT by introducing a metric-driven training process that improves both the quality and relevance of captions for autonomous driving. It uses a greedy search strategy to generate a baseline sequence, selecting the most probable token at each step, as shown in the following formula:

\begin{equation}
\hat{y}_t = \arg\max p(y_t | y_{1:t-1}, I).
\end{equation}

Here, \( \hat{y}_t \) represents the token chosen at time step \( t \), and \( p(y_t | y_{1:t-1}, I) \) denotes the probability of selecting token \( y_t \) given the previous tokens \( y_{1:t-1} \) and the input \( I \) (e.g., an image or a sequence of video frames). The sequence generated through this greedy approach serves as the baseline for reward calculation.

The essence of SCST lies in calculating rewards, which quantify the difference in CIDEr scores between a sampled sequence and its baseline. These token-specific rewards \( r_i \) are then employed to guide the training process, as defined by the following loss function:

\begin{equation}
\mathcal{L} = - \frac{1}{N} \sum_{i=1}^{N} \left( r_i \cdot \log p_i \cdot m_i \right).\nonumber
\end{equation}

where:

\begin{itemize}
    \item \( N \) is the number of non-padding tokens in the sequence.
    \item \( r_i \) is the reward associated with the \( i \)-th generated token.
    \item \( \log p_i \) is the logarithmic probability of generating the \( i \)-th token.
    \item \( m_i \) is a mask that excludes padding tokens (\( m_i = 1 \) for non-padding tokens, \( m_i = 0 \) for padding tokens).
\end{itemize}

The loss function maximize rewards by tuning the model parameters, thus encouraging the generation of sequences that produce higher evaluation scores.

We have made several optimizations to the implementation of SCST within the AVD2 system to differentiate it from the standard SCST framework:

\begin{itemize}
    \item \textbf{Masking and Loss Normalization:} To focus the model on meaningful content and minimize padding distortion, we use masking during reward calculation to exclude padding tokens. Additionally, we normalize the loss by the number of non-padding tokens, which helps stabilize training across sequences of varying lengths and prevents length-based bias~\cite{baevski2018adaptive,vaswani2017attention,ott2018scaling}.
    
    \item \textbf{Enhanced Focus on Contextual Relevance:} Within the AVD2 framework, SCST significantly improves the accuracy and contextual alignment of captions with visual content. This enhancement is crucial for capturing the nuanced details of complex driving scenarios and bolstering the reliability of autonomous driving systems.
\end{itemize}

\section{Experiment}
This section assesses the AVD2 framework using several key captioning metrics, including BLEU (B1, B2, B3, B4)~\cite{papineni2002bleu}, METEOR (M)~\cite{banerjee2005meteor}, ROUGE-L (R)~\cite{lin2004automatic}, Spice (S), CIDEr (C)~\cite{vedantam2015cider},  Fréchet Inception Distance (FID)~\cite{obukhov2020quality} and Video Inception Distance (VID)~\cite{heusel2017gans}. These metrics are vital for quantitatively measuring the quality of generated captions and videos. Additionally, ablation studies are performed to highlight the effectiveness of our system. We also compared our system's ability to understand accident scenarios with the popular large model, ChatGPT-4o, highlighting the strength of our model's comprehension capabilities.

\subsection{Implementation Details} 
In this experiment, the entire AVD2 framework training process was conducted on 4 NVIDIA Tesla V100 GPUs with a batch size of 4 per GPU, taking 60 hours to complete 49 epochs, and use several runs to calculate the average results. As for the Open-Sora 1.2 training process, both stages employed 8 NVIDIA A100 GPUs, with the first stage taking 35 hours for 11 epochs and the second stage taking 120 hours for 200 epochs. The generation of EMM-AU dataset took 16 NVIDIA A100 GPUs in 144 hours.

\subsection{Main Results} 
We evaluated the performance of our proposed AVD2 framework in comparison to the ADAPT framework using several standard captioning metrics. The evaluation was conducted on the original MM-AU dataset within the ADAPT framework and the MM-AU dataset within the AVD2 framework. Additionally, we present the results of applying the AVD2 framework to two data augmentation methods utilized in our research: the super-resolution (SR)~\cite{wang2018esrgan} augmented dataset and the EMM-AU dataset. We also compared SeViLA's performance on the Reasoning (Description) task with that of previous work ~\cite{fang2024abductive,brunet2011mathematical}. That work used SeViLA only to the
Reasoning task of the MM-AU dataset, which is a subset of the Description (Action+Reasoning) task in our paper. The outcomes for both the “description" task and the “avoidance" task are detailed in Table \uppercase\expandafter{\romannumeral1} and Table \uppercase\expandafter{\romannumeral2}, respectively. We also use the annotation of the original MM-AU dataset after designing as as the input prompt of ChatGPT-4o to lay down the style of generation with the enhancement of accidental comprehension. Table \uppercase\expandafter{\romannumeral3} shows the quality of pre-trained and fine-tuned Open-Sora generated videos.

\begin{table}[htbp]
    \centering
    \vspace{-1.0em}
    \caption{Evaluation Results of Description Task}
    \resizebox{\linewidth}{!}{
    \setlength{\tabcolsep}{3pt} 
    \begin{tabular}{c   c   c   c   c   c   c   c   c   c}
        \toprule
        Framework & Dataset & B1$\uparrow$ & B2$\uparrow$ & B3$\uparrow$ & B4$\uparrow$ & C$\uparrow$ & M$\uparrow$ & R$\uparrow$ & S$\uparrow$ \\
        \midrule
         SeViLA~\cite{fang2024abductive} & MM-AU & 27.0  & 21.6 & 18.1 &15.9  & 87.3 & 15.3 & 27.8 & 16.8 \\
        ChatGPT-4o & MM-AU & 10.4 & 5.38 & 1.27 & 0.31 & 7.26 & 15.8 & 12.2 & 5.14 \\
        ADAPT & MM-AU & 29.2 & 23.8 & 19.8 & 17.1 & 91.3 & 16.8 & 30.3 & 17.8 \\
        AVD2 & MM-AU & 30.2 & \textbf{24.3} & 20.2 & \textbf{17.9} & 97.6 & 18.0 & 30.8 & 18.6 \\
        AVD2 & SR & 30.1 & 24.0 & 20.1 & 17.6 & 89.3 & \textbf{18.4} & 30.1 & 18.3 \\
        AVD2 (Ours) & EMM-AU & \textbf{30.4}  & 24.3 & \textbf{20.3} & 17.8 & \textbf{98.1} & 17.2 & \textbf{31.2} & \textbf{18.9} \\
        \bottomrule
    \end{tabular}
    }
\end{table}

\begin{table}[htbp]
    \centering
    \vspace{-2.0em}
    \caption{Evaluation Results of Avoidance Task}
    \resizebox{\linewidth}{!}{
    \setlength{\tabcolsep}{3pt} 
    \begin{tabular}{c   c   c   c   c   c   c   c   c   c}
        \toprule
        Framework & Dataset & B1 $\uparrow$& B2$\uparrow$ & B3$\uparrow$ & B4$\uparrow$ & C$\uparrow$ & M$\uparrow$ & R $\uparrow$ & S$\uparrow$ \\
        \midrule
        ChatGPT-4o & MM-AU & 5.17 & 2.34 & 0.89 & 0.07 & 1.23 & 3.05 & 4.92 & 1.18\\
        ADAPT & MM-AU & 8.10 & 3.07 & 1.57 & 0.83 & 1.75 & \textbf{9.21} & 9.66 & 8.24 \\
        AVD2 & MM-AU & 9.90 & 3.83 & 1.82 & 0.90 & 2.44 & 9.11 & 10.14 & 8.58 \\
        AVD2 & SR & 8.90 & 3.29  & 1.75 & 0.87 & 2.35 & 9.18 & \textbf{10.3} & 8.57 \\
        AVD2 (Ours) & EMM-AU & \textbf{10.2} & \textbf{3.89} & \textbf{1.94} & \textbf{0.98} & \textbf{2.49} & 9.08 & 10.1 & \textbf{8.63} \\
        \bottomrule
    \end{tabular}
    }
\end{table}

\begin{table}[htbp]
    \centering
    \vspace{-2.0em}
    \caption{Video Before \& After Open-Sora Fine-tuning}
    \begin{tabular}{c  c   c}
        \toprule
        Framework & FID$\downarrow$ & VID $\downarrow$ \\
        \midrule
        Before & 149 &  332   \\
        After & \textbf{117} & \textbf{266}    \\
        \bottomrule
    \end{tabular}
\end{table}
\vspace{-1.2em}
The comparison demonstrates that the AVD2 framework outperforms the ADAPT framework, SeViLA model and ChatGPT-4o on the MM-AU dataset across all metrics in the “description" task and most metrics in the “avoidance" task, highlighting the effectiveness of our proposed improvements. While the overall metrics are modest due to the extended length of the MM-AU dataset's “avoidance" task, the AVD2 framework still enhances the performance of the ADAPT framework to a notable degree. Furthermore, when analyzing the evaluation metrics, the AVD2 framework shows superior results especially on the EMM-AU dataset generated when augmented with Open-Sora 1.2. The difference between the dataset enhanced with RRDBNet superscoring and the original dataset on the AVD2 is not significant, the superscoring enhancement does not have a significant enhancement effect on the AVD2 framework, and some of the evaluative metrics are even slightly lower than those of the original dataset, but all of them show a significant improvement on the performance of the original dataset compared to the performance of the original dataset on the ADAPT. As shown in the Table \uppercase\expandafter{\romannumeral3}, the evaluation metrics show better quality of the fine-tuned Open-Sora video.

\subsection{Ablation Study}
We demonstrate the effectiveness of our main contributions, including the AVD2-Understanding architecture and contributed dataset. The results of the “description" task and “avoidance" task are shown in Table \uppercase\expandafter{\romannumeral4} and Table \uppercase\expandafter{\romannumeral5}.

\begin{table}[htbp]
    \centering
    \vspace{-0.8em}
    \caption{Ablation Results of Description Task}
    \resizebox{\linewidth}{!}{
    \setlength{\tabcolsep}{3pt} 
    \begin{tabular}{c  c  c  c  c  c  c  c  c  c}
        \toprule
        Framework & Dataset & B1$\uparrow$ & B2$\uparrow$ & B3$\uparrow$ & B4$\uparrow$ & C$\uparrow$ & M$\uparrow$ & R$\uparrow$ & S$\uparrow$ \\
        \midrule
        ADAPT & SR & 28.2  & 22.3 & 18.5 & 16.0  & 88.3 & 16.6 & 29.6  & 17.6 \\
        AVD2 & SR & 30.1 & 24.0 & 20.1 & 17.6 & 89.3 & \textbf{18.4} & 30.1 & 18.3 \\
        ADAPT & EMM-AU & 29.8  & 23.9  & 20.1  & 17.5  & 95.5  & 16.9  & \textbf{31.2}  & 18.3  \\
        AVD2 (Ours) & EMM-AU & \textbf{30.4}  & \textbf{24.3} & \textbf{20.3} & \textbf{17.8} & \textbf{98.1} & 17.2 & 31.2 & \textbf{18.9}  \\
        \bottomrule
    \end{tabular}
    }
\end{table}

\begin{table}[htbp]
    \centering
    \vspace{-2.5em}
    \caption{Ablation Results of Avoidance Task}
    \resizebox{\linewidth}{!}{
    \setlength{\tabcolsep}{3pt} 
    \begin{tabular}{c  c  c  c  c  c  c  c  c  c}
        \toprule
        Framework & Dataset & B1 $\uparrow$ & B2 $\uparrow$ & B3$\uparrow$ & B4$\uparrow$ & C$\uparrow$ & M $\uparrow$& R$\uparrow$ & S$\uparrow$ \\
        \midrule
        ADAPT & SR & 9.61  & 3.45  & 1.61 & 0.81 & 2.10  & 9.15 & 9.99 & 8.56 \\
        AVD2 & SR & 9.10 & 3.29  & 1.75 & 0.87 & 2.35 & \textbf{9.18} & \textbf{10.3} & 8.57 \\
        ADAPT & EMM-AU & 9.98  & 3.80  & 1.86  & 0.94  & 2.41  & 9.14  & 10.1  & 8.61  \\
        AVD2 (Ours) & EMM-AU & \textbf{10.2} & \textbf{3.89} & \textbf{1.94} & \textbf{0.98} & \textbf{2.49} & 9.08 & 10.1 & \textbf{8.63}  \\
        \bottomrule
    \end{tabular}
    }
\end{table}
\vspace{-1.0em}
Combining the above evaluation indicators, the AVD2 framework outperforms the ADAPT framework in the vast majority of cases in both of the tasks, proving the effectiveness of the AVD2 framework.

\subsection{Manual Assessment}
We also conducted a manual evaluation of the test set results from the AVD2 and ADAPT frameworks of the Description and Avoidance task, calculating accuracy and comparing these results with those of the SeViLA model from previous work  ~\cite{fang2024abductive} and ChatGPT-4o~\cite{hwang2024safe}. The results of the manual evaluation are shown in the Table \uppercase\expandafter{\romannumeral6}, which also shows the best performance of AVD2 framework.

\begin{table}[htbp]
    \centering
    \vspace{-1.0em}
    \caption{Results of Manual Assessment}
    \begin{tabular}{c  c  c  c}
        \toprule
        Task & Framework & Dataset & Accuracy$\uparrow$ \\
        \midrule
        \multirow{4}{*}{Description}
        & SeViLA~\cite{fang2024abductive} & MM-AU & 89.02  \\
        & ChatGPT-4o & MM-AU & 55.00    \\
        & ADAPT & MM-AU & 87.98  \\
        & AVD2 (Ours) & EMM-AU & \textbf{90.12} \\
        \midrule
        \multirow{3}{*}{Avoidance}
        & ChatGPT-4o & MM-AU & 52.50     \\
        & ADAPT & MM-AU & 75.47   \\
        & AVD2 (Ours) & EMM-AU & \textbf{77.49} \\
        \bottomrule
    \end{tabular}
\end{table}
\vspace{-0.8em}
In this section we visualize AVD2 in terms of its ability to understand accident videos. Two original MM-AU videos are also randomly selected to compare the ability of the ChatGPT-4o with AVD2 to understand video incidents. We also input some original annotations of MM-AU dataset as the ChatGPT-4o's prompt, to allow it better understanding the task and lay down the output style.

\begin{figure}[htbp]
\centering
\includegraphics[width=0.8\linewidth]{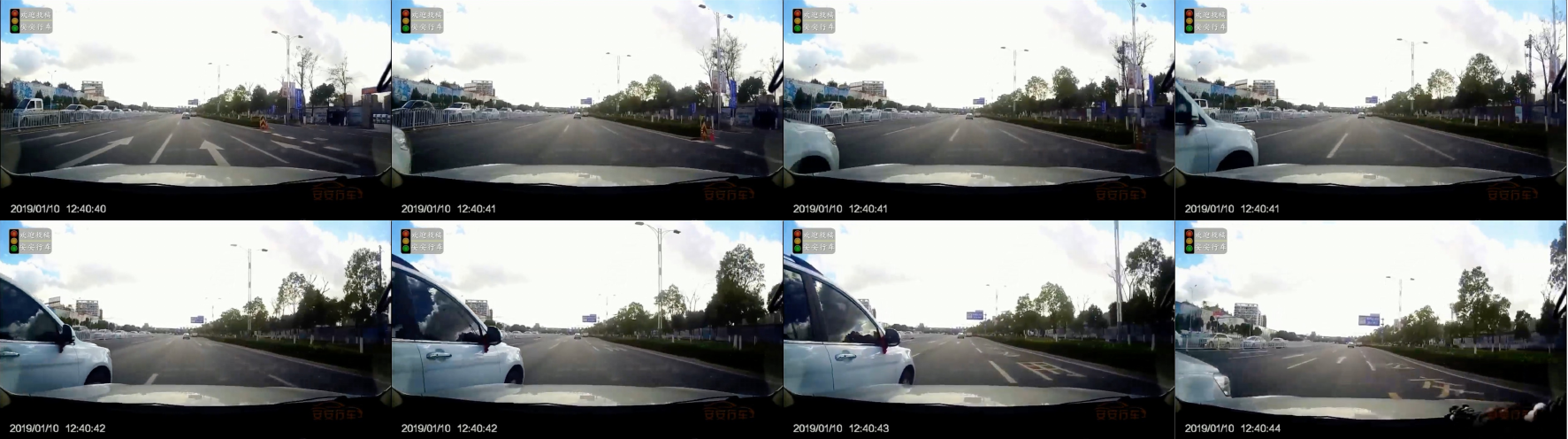} 
\caption{\textbf{Visualization for the First Video Understanding}}
\label{fig:first_accident}
\vspace{-2.0em}
\end{figure}

\begin{figure}[htbp]
\centering
\includegraphics[width=0.8\linewidth]{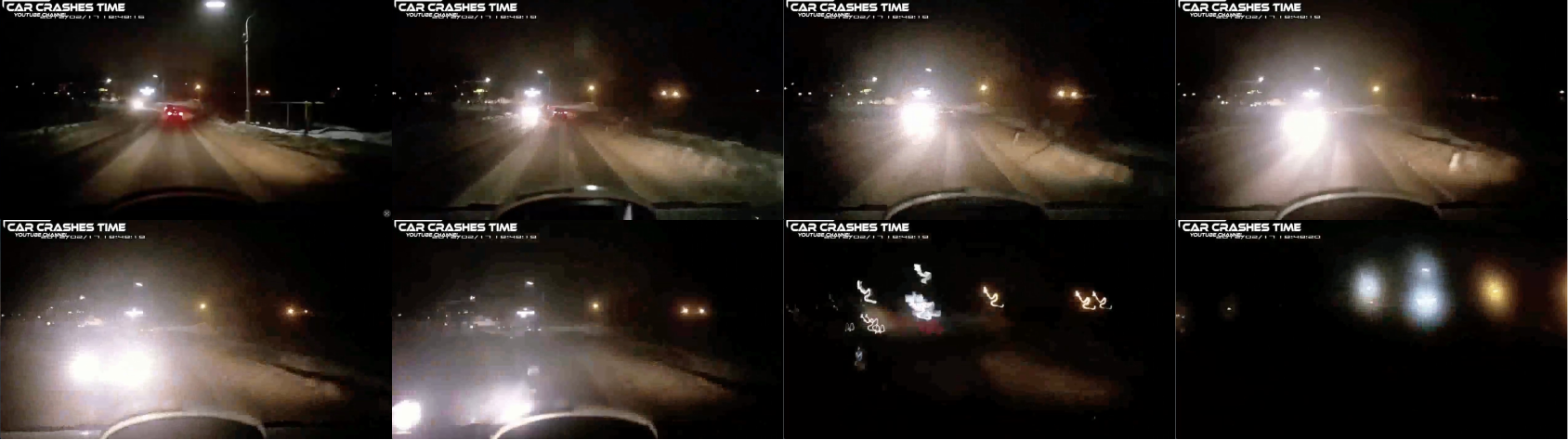} 
\caption{\textbf{Visualization for the Second Video Understanding}}
\label{fig:second_accident}
\end{figure}

The visualization of accident understanding results are shown in the Fig.~\ref{fig:first_accident} and \ref{fig:second_accident}. In these figures, “D” stands for “description”, “A” stands for “avoidance”, and the two example videos are selected to be placed. After comparative analysis, the comprehension of the AVD2 framework essentially matches that of Ground Truth and is consistent with the general idea.

{   \scriptsize

    \noindent\textbf{AVD2 Prediction:}\\
    \textcolor{red}{D:}
    \hspace*{0.3em}\begin{minipage}[t]{\dimexpr\linewidth-1.5em}
    A vehicle changes lanes with the same direction to ego-car; Vehicles don't give way to normal driving vehicles when turning or changing lanes.\\
    \end{minipage}
    
    \noindent\textcolor{blue}{A:}
    \hspace*{0.3em}\begin{minipage}[t]{\dimexpr\linewidth-1.5em}
    Before turning or changing lanes, vehicles should turn on the turn signal in advance, observe the surrounding vehicles and control the speed. When driving, vehicles should abide by traffic rules, and give the way for the normal running vehicles. Vehicles that will enter the main road should give way to the vehicles which drive on the main road or leave the main road. Vehicles that drive on the auxiliary road should give way to the vehicles which drive off the main road.\\
    \end{minipage}

    \noindent\textbf{ChatGPT-4o Prediction:}\\
    \textcolor{red}{D:}
    \hspace*{0.3em}\begin{minipage}[t]{\dimexpr\linewidth-1.5em}
    A vehicle approaches a busy intersection and fails to notice another car coming from the side; The vehicle abruptly brakes to avoid a collision, but the close proximity creates a dangerous situation.\\
    \end{minipage}
    
    \noindent\textcolor{blue}{A:}
    \hspace*{0.3em}\begin{minipage}[t]{\dimexpr\linewidth-1.5em}
    Drivers should always reduce speed when approaching intersections and remain alert to traffic from all directions. Maintaining a safe distance and carefully observing other vehicles is essential to prevent accidents at intersections.\\
    \end{minipage}
    
    \noindent\textbf{GroundTruth:}\\
    \textcolor{red}{D:}
    \hspace*{0.3em}\begin{minipage}[t]{\dimexpr\linewidth-1.5em}
    Lead vehicle stops; Vehicles do not give way to normal driving vehicles when turning or changing lanes.\\
    \end{minipage}
    
    \noindent\textcolor{blue}{A:}
    \hspace*{0.3em}\begin{minipage}[t]{\dimexpr\linewidth-1.5em}
    Before turning or changing lanes, vehicles should turn on the turn signal in advance, observe the surrounding vehicles and control the speed. When driving, vehicles should abide by traffic rules, and give the way for the normal running vehicles. Vehicles that will enter the main road should give way to the vehicles which drive on the main road or leave the main road. Vehicles that drive on the auxiliary road should give way to the vehicles which drive off the main road.\\
    \end{minipage}

} 

\vspace{-0.5em}
{   \scriptsize
    \noindent\textbf{AVD2 Prediction:}\\
    \textcolor{red}{D:}
    \hspace*{0.3em}\begin{minipage}[t]{\dimexpr\linewidth-1.5em}
    A vehicle changes lanes with the same direction to ego-car / another vehicle; Ego-car drives too fast and the braking distance is short. \\
    \end{minipage}
    
    \noindent\textcolor{blue}{A:}
    \hspace*{0.3em}\begin{minipage}[t]{\dimexpr\linewidth-1.5em}
    Ego-cars should not exceed the speed limit during driving, slow down when passing intersections or crosswalks, especially for areas with many pedestrians.\\
    \end{minipage}

    \noindent\textbf{ChatGPT-4o Prediction:}\\
    \textcolor{red}{D:}
    \hspace*{0.3em}\begin{minipage}[t]{\dimexpr\linewidth-1.5em}
    A vehicle makes a sharp turn at an intersection without signaling; The vehicle behind is forced to brake abruptly due to insufficient reaction time. \\
    \end{minipage}
    
    \noindent\textcolor{blue}{A:}
    \hspace*{0.3em}\begin{minipage}[t]{\dimexpr\linewidth-1.5em}
    Drivers should signal well in advance before making turns at intersections. Maintaining a safe distance from other vehicles and anticipating sudden turns can help prevent accidents.\\
    \end{minipage}
    
    \noindent\textbf{GroundTruth:}\\
    \textcolor{red}{D:}
    \hspace*{0.3em}\begin{minipage}[t]{\dimexpr\linewidth-1.5em}
    Vehicles meet on the road; Vehicles drive too fast with short braking distance.\\
    \end{minipage}
    
    \noindent\textcolor{blue}{A:}
    \hspace*{0.3em}\begin{minipage}[t]{\dimexpr\linewidth-1.5em}
    Vehicles should not exceed the speed limit during driving, especially in areas with many pedestrians. Vehicles should slow down when passing intersections or crosswalks, and observe the traffic carefully.
    \end{minipage}
}

\section{Conclusion}
In this study, we address traffic accident understanding in autonomous driving using the AVD2 framework, which generates detailed descriptions, identifies causes, and suggests preventive measures. We also introduce the EMM-AU dataset, pioneering the generation of accident videos. Results demonstrate that AVD2 enhances accident analysis and prevention, establishing a new benchmark for safety in autonomous driving systems.





\clearpage 



\begin{thebibliography}{99}

\bibitem{chen2024end}
Li Chen, Penghao Wu, Kashyap Chitta, Bernhard Jaeger, Andreas Geiger, and Hongyang Li, "End-to-end autonomous driving: Challenges and frontiers," \textit{IEEE Transactions on Pattern Analysis and Machine Intelligence}, 2024, IEEE.

\bibitem{korhonen2012peak}
Jari Korhonen and Junyong You, "Peak signal-to-noise ratio revisited: Is simple beautiful?" in \textit{2012 Fourth International Workshop on Quality of Multimedia Experience}, 2012, pp. 37--38, IEEE.

\bibitem{jin2023adapt}
Bu Jin, Xinyu Liu, Yupeng Zheng, Pengfei Li, Hao Zhao, Tong Zhang, Yuhang Zheng, Guyue Zhou, and Jingjing Liu, "Adapt: Action-aware driving caption transformer," in \textit{2023 IEEE International Conference on Robotics and Automation (ICRA)}, 2023, pp. 7554--7561, IEEE.

\bibitem{marcu2024lingoqa}
Ana-Maria Marcu, Long Chen, Jan H{\"u}nermann, Alice Karnsund, Benoit Hanotte, Prajwal Chidananda, Saurabh Nair, Vijay Badrinarayanan, Alex Kendall, Jamie Shotton, and others, "Lingoqa: Visual question answering for autonomous driving," in \textit{European Conference on Computer Vision}, 2024, pp. 252--269, Springer.

\bibitem{atakishiyev2023explaining}
Shahin Atakishiyev, Mohammad Salameh, Housam Babiker, and Randy Goebel, "Explaining autonomous driving actions with visual question answering," in \textit{2023 IEEE 26th International Conference on Intelligent Transportation Systems (ITSC)}, 2023, pp. 1207--1214, IEEE.

\bibitem{xu2021sutd}
Li Xu, He Huang, and Jun Liu, "Sutd-trafficqa: A question answering benchmark and an efficient network for video reasoning over traffic events," in \textit{Proceedings of the IEEE/CVF Conference on Computer Vision and Pattern Recognition}, 2021, pp. 9878--9888.

\bibitem{qian2024nuscenes}
Tianwen Qian, Jingjing Chen, Linhai Zhuo, Yang Jiao, and Yu-Gang Jiang, "Nuscenes-qa: A multi-modal visual question answering benchmark for autonomous driving scenario," in \textit{Proceedings of the AAAI Conference on Artificial Intelligence}, vol. 38, no. 5, pp. 4542--4550, 2024.

\bibitem{le2020hierarchical}
Thao Minh Le, Vuong Le, Svetha Venkatesh, and Truyen Tran, "Hierarchical conditional relation networks for video question answering," in \textit{Proceedings of the IEEE/CVF Conference on Computer Vision and Pattern Recognition}, 2020, pp. 9972--9981.

\bibitem{kim2017interpretable}
Jinkyu Kim and John Canny, "Interpretable learning for self-driving cars by visualizing causal attention," in \textit{Proceedings of the IEEE International Conference on Computer Vision}, 2017, pp. 2942--2950.

\bibitem{zeng2019end}
Wenyuan Zeng, Wenjie Luo, Simon Suo, Abbas Sadat, Bin Yang, Sergio Casas, and Raquel Urtasun, "End-to-end interpretable neural motion planner," in \textit{Proceedings of the IEEE/CVF Conference on Computer Vision and Pattern Recognition}, 2019, pp. 8660--8669.

\bibitem{kim2018textual}
Jinkyu Kim, Anna Rohrbach, Trevor Darrell, John Canny, and Zeynep Akata, "Textual explanations for self-driving vehicles," in \textit{Proceedings of the European Conference on Computer Vision (ECCV)}, 2018, pp. 563--578.

\bibitem{bujimalla2020b}
Shashank Bujimalla, Mahesh Subedar, and Omesh Tickoo, "B-SCST: Bayesian self-critical sequence training for image captioning," \textit{arXiv preprint arXiv:2004.02435}, 2020.

\bibitem{liu2023cross}
Yang Liu, Guanbin Li, and Liang Lin, "Cross-modal causal relational reasoning for event-level visual question answering," \textit{IEEE Transactions on Pattern Analysis and Machine Intelligence}, vol. 45, no. 10, pp. 11624--11641, 2023, IEEE.

\bibitem{nahata2021assessing}
Richa Nahata, Daniel Omeiza, Rhys Howard, and Lars Kunze, "Assessing and explaining collision risk in dynamic environments for autonomous driving safety," in \textit{2021 IEEE International Intelligent Transportation Systems Conference (ITSC)}, 2021, pp. 223--230, IEEE.

\bibitem{hwang2024safe}
Hochul Hwang, Sunjae Kwon, Yekyung Kim, and Donghyun Kim, "Is it safe to cross? Interpretable Risk Assessment with GPT-4V for Safety-Aware Street Crossing," \textit{arXiv preprint arXiv:2402.06794}, 2024.

\bibitem{ding2024hint}
Kairui Ding, Boyuan Chen, Yuchen Su, Huan-ang Gao, Bu Jin, Chonghao Sima, Wuqiang Zhang, Xiaohui Li, Paul Barsch, Hongyang Li, and others, "Hint-ad: Holistically aligned interpretability in end-to-end autonomous driving," \textit{arXiv preprint arXiv:2409.06702}, 2024.

\bibitem{li2023understanding}
Yang Li, Xiaoxue Chen, Hao Zhao, Jiangtao Gong, Guyue Zhou, Federico Rossano, and Yixin Zhu, "Understanding Embodied Reference with Touch-Line Transformer," in \textit{ICLR}, 2023.

\bibitem{inoue2023towards}
Naoto Inoue, Kotaro Kikuchi, Edgar Simo-Serra, Mayu Otani, and Kota Yamaguchi, "Towards flexible multi-modal document models," in \textit{Proceedings of the IEEE/CVF Conference on Computer Vision and Pattern Recognition}, 2023, pp. 14287--14296.

\bibitem{venugopalan2015sequence}
Subhashini Venugopalan, Marcus Rohrbach, Jeffrey Donahue, Raymond Mooney, Trevor Darrell, and Kate Saenko, "Sequence to sequence-video to text," in \textit{Proceedings of the IEEE International Conference on Computer Vision}, 2015, pp. 4534--4542.

\bibitem{gao2023retrieval}
Yunfan Gao, Yun Xiong, Xinyu Gao, Kangxiang Jia, Jinliu Pan, Yuxi Bi, Yi Dai, Jiawei Sun, and Haofen Wang, "Retrieval-augmented generation for large language models: A survey," \textit{arXiv preprint arXiv:2312.10997}, 2023.

\bibitem{wang2021end}
Teng Wang, Ruimao Zhang, Zhichao Lu, Feng Zheng, Ran Cheng, and Ping Luo, "End-to-end dense video captioning with parallel decoding," in \textit{Proceedings of the IEEE/CVF International Conference on Computer Vision}, 2021, pp. 6847--6857.

\bibitem{lin2022swinbert}
Kevin Lin, Linjie Li, Chung-Ching Lin, Faisal Ahmed, Zhe Gan, Zicheng Liu, Yumao Lu, and Lijuan Wang, "Swinbert: End-to-end transformers with sparse attention for video captioning," in \textit{Proceedings of the IEEE/CVF Conference on Computer Vision and Pattern Recognition}, 2022, pp. 17949--17958.

\bibitem{liu2023delving}
Xinyu Liu, Beiwen Tian, Zhen Wang, Rui Wang, Kehua Sheng, Bo Zhang, Hao Zhao, and Guyue Zhou, "Delving into shape-aware zero-shot semantic segmentation," in \textit{Proceedings of the IEEE/CVF Conference on Computer Vision and Pattern Recognition}, 2023, pp. 2999--3009.
\bibitem{zheng2024monoocc}

Yupeng Zheng, Xiang Li, Pengfei Li, Yuhang Zheng, Bu Jin, Chengliang Zhong, Xiaoxiao Long, Hao Zhao, and Qichao Zhang, "Monoocc: Digging into monocular semantic occupancy prediction," \textit{arXiv preprint arXiv:2403.08766}, 2024.

\bibitem{vondrick2016generating}
Carl Vondrick, Hamed Pirsiavash, and Antonio Torralba, "Generating videos with scene dynamics," \textit{Advances in Neural Information Processing Systems}, vol. 29, 2016.

\bibitem{cai2018deep}
Haoye Cai, Chunyan Bai, Yu-Wing Tai, and Chi-Keung Tang, "Deep video generation, prediction and completion of human action sequences," in \textit{Proceedings of the European Conference on Computer Vision (ECCV)}, 2018, pp. 366--382.

\bibitem{huang2024vbench}
Ziqi Huang, Yinan He, Jiashuo Yu, Fan Zhang, Chenyang Si, Yuming Jiang, Yuanhan Zhang, Tianxing Wu, Qingyang Jin, Nattapol Chanpaisit, and others, "Vbench: Comprehensive benchmark suite for video generative models," in \textit{Proceedings of the IEEE/CVF Conference on Computer Vision and Pattern Recognition}, 2024, pp. 21807--21818.

\bibitem{chen2023pixart}
Junsong Chen, Jincheng Yu, Chongjian Ge, Lewei Yao, Enze Xie, Yue Wu, Zhongdao Wang, James Kwok, Ping Luo, Huchuan Lu, and others, "PixArt-{$\alpha$}: Fast Training of Diffusion Transformer for Photorealistic Text-to-Image Synthesis," \textit{arXiv preprint arXiv:2310.00426}, 2023.

\bibitem{hussain2017automatic}
Zaeem Hussain, Mingda Zhang, Xiaozhong Zhang, Keren Ye, Christopher Thomas, Zuha Agha, Nathan Ong, and Adriana Kovashka, "Automatic understanding of image and video advertisements," in \textit{Proceedings of the IEEE Conference on Computer Vision and Pattern Recognition}, 2017, pp. 1705--1715.

\bibitem{karras2023dreampose}
Johanna Karras, Aleksander Holynski, Ting-Chun Wang, and Ira Kemelmacher-Shlizerman, "Dreampose: Fashion image-to-video synthesis via stable diffusion," in \textit{2023 IEEE/CVF International Conference on Computer Vision (ICCV)}, 2023, pp. 22623--22633, IEEE.

\bibitem{jiang2023text2performer}
Yuming Jiang, Shuai Yang, Tong Liang Koh, Wayne Wu, Chen Change Loy, and Ziwei Liu, "Text2performer: Text-driven human video generation," in \textit{Proceedings of the IEEE/CVF International Conference on Computer Vision}, 2023, pp. 22747--22757.

\bibitem{wen2024panacea}
Yuqing Wen, Yucheng Zhao, Yingfei Liu, Fan Jia, Yanhui Wang, Chong Luo, Chi Zhang, Tiancai Wang, Xiaoyan Sun, and Xiangyu Zhang, "Panacea: Panoramic and controllable video generation for autonomous driving," in \textit{Proceedings of the IEEE/CVF Conference on Computer Vision and Pattern Recognition}, 2024, pp. 6902--6912.

\bibitem{yuan2024magictime}
Shenghai Yuan, Jinfa Huang, Yujun Shi, Yongqi Xu, Ruijie Zhu, Bin Lin, Xinhua Cheng, Li Yuan, and Jiebo Luo, "MagicTime: Time-lapse Video Generation Models as Metamorphic Simulators," \textit{arXiv preprint arXiv:2404.05014}, 2024.

\bibitem{zhang2024ctrl}
Guiyu Zhang, Huan-ang Gao, Zijian Jiang, Hao Zhao, and Zhedong Zheng, "Ctrl-u: Robust conditional image generation via uncertainty-aware reward modeling," \textit{arXiv preprint arXiv:2410.11236}, 2024.

\bibitem{xu2024diffusion}
Zhiyuan Xu, Yinhe Chen, Huan-ang Gao, Weiyan Zhao, Guiyu Zhang, and Hao Zhao, "Diffusion-based Visual Anagram as Multi-task Learning," \textit{arXiv preprint arXiv:2412.02693}, 2024.

\bibitem{gao2024scp}
Huan-ang Gao, Mingju Gao, Jiaju Li, Wenyi Li, Rong Zhi, Hao Tang, and Hao Zhao, "SCP-Diff: Spatial-Categorical Joint Prior for Diffusion Based Semantic Image Synthesis," in \textit{European Conference on Computer Vision}, 2024, pp. 37--54, Springer.

\bibitem{li2024fairdiff}
Wenyi Li, Haoran Xu, Guiyu Zhang, Huan-ang Gao, Mingju Gao, Mengyu Wang, and Hao Zhao, "Fairdiff: Fair segmentation with point-image diffusion," in \textit{International Conference on Medical Image Computing and Computer-Assisted Intervention}, 2024, pp. 617--628, Springer.

\bibitem{gu2025text2street}
Songen Gu, Jinming Su, Yiting Duan, Xingyue Chen, Junfeng Luo, and Hao Zhao, "Text2street: Controllable text-to-image generation for street views," in \textit{International Conference on Pattern Recognition}, 2025, pp. 130--145, Springer.

\bibitem{chang2024llmscenario}
Cheng Chang, Siqi Wang, Jiawei Zhang, Jingwei Ge, and Li Li, "LLMScenario: Large Language Model Driven Scenario Generation," \textit{IEEE Transactions on Systems, Man, and Cybernetics: Systems}, 2024, IEEE.

\bibitem{tulyakov2018mocogan}
Sergey Tulyakov, Ming-Yu Liu, Xiaodong Yang, and Jan Kautz, "Mocogan: Decomposing motion and content for video generation," in \textit{Proceedings of the IEEE Conference on Computer Vision and Pattern Recognition}, 2018, pp. 1526--1535.

\bibitem{pang2021image}
Yingxue Pang, Jianxin Lin, Tao Qin, and Zhibo Chen, "Image-to-image translation: Methods and applications," \textit{IEEE Transactions on Multimedia}, vol. 24, pp. 3859--3881, 2021, IEEE.

\bibitem{denton2018stochastic}
Emily Denton and Rob Fergus, "Stochastic video generation with a learned prior," in \textit{International Conference on Machine Learning}, 2018, pp. 1174--1183, PMLR.

\bibitem{gupta2018social}
Agrim Gupta, Justin Johnson, Li Fei-Fei, Silvio Savarese, and Alexandre Alahi, "Social GAN: Socially acceptable trajectories with generative adversarial networks," in \textit{Proceedings of the IEEE Conference on Computer Vision and Pattern Recognition}, 2018, pp. 2255--2264.

\bibitem{fang2024abductive}
Jianwu Fang, Lei-lei Li, Junfei Zhou, Junbin Xiao, Hongkai Yu, Chen Lv, Jianru Xue, and Tat-Seng Chua, "Abductive Ego-View Accident Video Understanding for Safe Driving Perception," in \textit{Proceedings of the IEEE/CVF Conference on Computer Vision and Pattern Recognition}, 2024, pp. 22030--22040.

\bibitem{zheng2024open}
Zangwei Zheng, Xiangyu Peng, Tianji Yang, Chenhui Shen, Shenggui Li, Hongxin Liu, Yukun Zhou, Tianyi Li, and Yang You, "Open-sora: Democratizing efficient video production for all," \textit{arXiv preprint arXiv:2412.20404}, 2024.

\bibitem{blattmann2023stable}
Andreas Blattmann, Tim Dockhorn, Sumith Kulal, Daniel Mendelevitch, Maciej Kilian, Dominik Lorenz, Yam Levi, Zion English, Vikram Voleti, Adam Letts, and others, "Stable video diffusion: Scaling latent video diffusion models to large datasets," \textit{arXiv preprint arXiv:2311.15127}, 2023.

\bibitem{wang2021real}
Xintao Wang, Liangbin Xie, Chao Dong, and Ying Shan, "Real-ESRGAN: Training real-world blind super-resolution with pure synthetic data," in \textit{Proceedings of the IEEE/CVF International Conference on Computer Vision}, 2021, pp. 1905--1914.

\bibitem{ads_of_the_world}
\textit{Ads of the World}, No Author, n.d., Available at: \url{https://www.adsoftheworld.com/} [Accessed Apr. 12, 2024].

\bibitem{cannes_lions}
Cannes, \textit{Cannes Lions}, 2017, Available at: \url{https://www.canneslions.com/} [Accessed Apr. 23, 2024].

\bibitem{brooks2024video}
Tim Brooks, Bill Peebles, Connor Holmes, Will DePue, Yufei Guo, Li Jing, David Schnurr, Joe Taylor, Troy Luhman, Eric Luhman, and others, "Video generation models as world simulators," \textit{arXiv preprint arXiv:2410.11236}, 2024, Available at: \url{https://openai.com/research/video-generation-models-as-world-simulators}.

\bibitem{kang2023scaling}
Minguk Kang, Jun-Yan Zhu, Richard Zhang, Jaesik Park, Eli Shechtman, Sylvain Paris, and Taesung Park, "Scaling up GANs for text-to-image synthesis," in \textit{Proceedings of the IEEE/CVF Conference on Computer Vision and Pattern Recognition}, 2023, pp. 10124--10134.

\bibitem{wang2018esrgan}
Xintao Wang, Ke Yu, Shixiang Wu, Jinjin Gu, Yihao Liu, Chao Dong, Yu Qiao, and Chen Change Loy, "ESRGAN: Enhanced super-resolution generative adversarial networks," in \textit{Proceedings of the European Conference on Computer Vision (ECCV) Workshops}, 2018.

\bibitem{dosovitskiy2016generating}
Alexey Dosovitskiy and Thomas Brox, "Generating images with perceptual similarity metrics based on deep networks," \textit{Advances in Neural Information Processing Systems}, vol. 29, 2016.

\bibitem{zhang2023blind}
Chongqi Zhang, Ziwen Zhang, Yao Deng, Yueyi Zhang, Mingzhe Chong, Yunhua Tan, and Pukun Liu, "Blind super-resolution for SAR images with speckle noise based on deep learning probabilistic degradation model and SAR priors," \textit{Remote Sensing}, vol. 15, no. 2, pp. 330, 2023, MDPI.

\bibitem{yu2023language}
Lijun Yu, José Lezama, Nitesh B Gundavarapu, Luca Versari, Kihyuk Sohn, David Minnen, Yong Cheng, Vighnesh Birodkar, Agrim Gupta, Xiuye Gu, and others, "Language Model Beats Diffusion--Tokenizer is Key to Visual Generation," \textit{arXiv preprint arXiv:2310.05737}, 2023.

\bibitem{li2024carbon}
Baolin Li, Yankai Jiang, and Devesh Tiwari, "Carbon in Motion: Characterizing Open-Sora on the Sustainability of Generative AI for Video Generation," in \textit{3rd Workshop on Sustainable Computer Systems (HotCarbon)}, 2024.

\bibitem{rennie2017self}
Steven J Rennie, Etienne Marcheret, Youssef Mroueh, Jerret Ross, and Vaibhava Goel, "Self-critical sequence training for image captioning," in \textit{Proceedings of the IEEE Conference on Computer Vision and Pattern Recognition}, 2017, pp. 7008--7024.

\bibitem{xu2019multi}
Ning Xu, Hanwang Zhang, An-An Liu, Weizhi Nie, Yuting Su, Jie Nie, and Yongdong Zhang, "Multi-level policy and reward-based deep reinforcement learning framework for image captioning," \textit{IEEE Transactions on Multimedia}, vol. 22, no. 5, pp. 1372--1383, 2019, IEEE.

\bibitem{luo2018discriminability}
Ruotian Luo, Brian Price, Scott Cohen, and Gregory Shakhnarovich, "Discriminability objective for training descriptive captions," in \textit{Proceedings of the IEEE Conference on Computer Vision and Pattern Recognition}, 2018, pp. 6964--6974.

\bibitem{luo2020better}
Ruotian Luo, "A better variant of self-critical sequence training," \textit{arXiv preprint arXiv:2003.09971}, 2020.

\bibitem{baevski2018adaptive}
Alexei Baevski and Michael Auli, "Adaptive input representations for neural language modeling," \textit{arXiv preprint arXiv:1809.10853}, 2018.

\bibitem{vaswani2017attention}
A Vaswani, "Attention is all you need," \textit{Advances in Neural Information Processing Systems}, 2017.

\bibitem{ott2018scaling}
Myle Ott, Sergey Edunov, David Grangier, and Michael Auli, "Scaling neural machine translation," \textit{arXiv preprint arXiv:1806.00187}, 2018.

\bibitem{papineni2002bleu}
Kishore Papineni, Salim Roukos, Todd Ward, and Wei-Jing Zhu, "BLEU: A method for automatic evaluation of machine translation," in \textit{Proceedings of the 40th Annual Meeting of the Association for Computational Linguistics}, 2002, pp. 311--318.

\bibitem{banerjee2005meteor}
Satanjeev Banerjee and Alon Lavie, "METEOR: An automatic metric for MT evaluation with improved correlation with human judgments," in \textit{Proceedings of the ACL Workshop on Intrinsic and Extrinsic Evaluation Measures for Machine Translation and/or Summarization}, 2005, pp. 65--72.

\bibitem{lin2004automatic}
Chin-Yew Lin and Franz Josef Och, "Automatic evaluation of machine translation quality using longest common subsequence and skip-bigram statistics," in \textit{Proceedings of the 42nd Annual Meeting of the Association for Computational Linguistics (ACL-04)}, 2004, pp. 605--612.

\bibitem{vedantam2015cider}
Ramakrishna Vedantam, C Lawrence Zitnick, and Devi Parikh, "CIDER: Consensus-based image description evaluation," in \textit{Proceedings of the IEEE Conference on Computer Vision and Pattern Recognition}, 2015, pp. 4566--4575.

\bibitem{obukhov2020quality}
Artem Obukhov and Mikhail Krasnyanskiy, "Quality assessment method for GAN based on modified metrics inception score and Fréchet inception distance," in \textit{Software Engineering Perspectives in Intelligent Systems: Proceedings of 4th Computational Methods in Systems and Software 2020, Vol. 1}, 2020, pp. 102--114, Springer.

\bibitem{heusel2017gans}
Martin Heusel, Hubert Ramsauer, Thomas Unterthiner, Bernhard Nessler, and Sepp Hochreiter, "GANs trained by a two time-scale update rule converge to a local Nash equilibrium," \textit{Advances in Neural Information Processing Systems}, vol. 30, 2017.

\bibitem{brunet2011mathematical}
Dominique Brunet, Edward R Vrscay, and Zhou Wang, "On the mathematical properties of the structural similarity index," \textit{IEEE Transactions on Image Processing}, vol. 21, no. 4, pp. 1488--1499, 2011, IEEE.








































\end{thebibliography}
\end{document}